\documentclass{article}

\usepackage{PRIMEarxiv}

\usepackage[utf8]{inputenc} % allow utf-8 input
\usepackage[T1]{fontenc}    % use 8-bit T1 fonts
\usepackage{hyperref}       % hyperlinks
\usepackage{url}            % simple URL typesetting
\usepackage{booktabs}       % professional-quality tables
\usepackage{amsfonts}       % blackboard math symbols
\usepackage{nicefrac}       % compact symbols for 1/2, etc.
\usepackage{microtype}      % microtypography
\usepackage{lipsum}
\usepackage{fancyhdr}       % header
\usepackage{graphicx}       % graphics
\graphicspath{{media/}}     % organize your images and other figures under media/ folder

\title{Anthropomorphic finger for grasping applications: 3D printed endoskeleton in a soft skin
\thanks{\textit{\underline{Citation}}: 
\textbf{Tavakoli, M., Sayuk, A., Lourenço, J. et al. Anthropomorphic finger for grasping applications: 3D printed endoskeleton in a soft skin. Int J Adv Manuf Technol 91, 2607–2620 (2017). DOI: 10.1007/s00170-016-9971-8}} 
}

\author{
  Mahmoud Tavakoli, Andriy Sayuk, Jo\~ao Louren\c{c}o, Pedro Neto \\
  University of Coimbra \\
  Coimbra\\
}

\begin{document}
\maketitle

\begin{abstract}
Application of soft and compliant joints in grasping mechanisms received an increasing attention during recent years. This article suggests the design and development of a novel bio-inspired compliant finger which is composed of a 3D printed rigid endoskeleton covered by a soft matter. The overall integrated system resembles a biological structure in which a finger presents an anthropomorphic look. The mechanical properties of such structure are enhanced through optimization of the repetitive geometrical structures that constructs a flexure bearing as a joint for the fingers. The endoskeleton is formed by additive manufacturing of such geometries with rigid materials. The geometry of the endoskeleton was studied by Finite Element Analysis (FEA) to obtain the desired properties: high stiffness against lateral deflection and twisting, and low stiffness in the desired bending axis of the fingers. Results are validated by experimental analysis.
\end{abstract}

% keywords can be removed

\section{Introduction}
\label{intro}
Integration of compliance in mechanical systems in general and in robotic systems in particular received an increasing attention in the robotics community during the recent years. For the design and manufacturing of robotic hands, compliance can result in a passive adaptability in articulated fingers of a grasping mechanism \cite{Tavakoli201623} \cite{Schultz2014}. In addition, integration of the compliance enhances the safety in the human-robot interaction process. An important subject of research in this domain is the design and development of joints with a compliant behaviour, which can be formed by a proper controller. The DLR HAND II \cite{butterfass2001dlr} integrates the compliance into the control loop by using an appropriate impedance controller. Yet, this solution demands for a sophisticated controller with a fast response time. The development of grippers made of soft materials has been in studied in the last few years, including the design of anthropomorphic fingers \cite{Rateni2015} \cite{Cianchetti15}.

Compliance can also be integrated in the actuator.  Meka-H2 compliant hand ~\cite{MekaBot2009} integrates a Series Elastic Actuator (SEA). It has a total of 12 Degrees Of Freedom (DOF) controlled by 5 SEA actuators. Another approach is to directly integrate compliance into the joints. Pisa-IIT Softhand \cite{godfrey2013SoftHand}, and the UB hand \cite{melchiorri2013-UBhand} are examples of recent development of anthropomorphic hands that directly integrate the compliance into the joints.  In Pisa-IIT Softhand several elements are interconnected with elastic elements, while in UB hand the joint compliance is achieved by integration of springs into the joints. Flexirigid \cite{Tavakoli13-1}, SDM hand \cite{dollar2010}, and ISR-Softhand \cite{Tavakoli2014} integrate elastomeric joints between their rigid phalanges. The elastic joints of the SDM hand \cite{dollar2010}, are formed by casting a urethane rubber into the moulds that are built in the 3D printed finger. The shape deposition manufacturing originally reported in \cite{dollar2006robust} was later updated with the Hybrid Deposition Manufacturing (HDM) \cite{ma2015hybrid}, in which the multipart moulds can be reused instead of destroying the temporary mould features, making this method even less labour intensive. Another approach on development of the Softhand was presented in RBO hands \cite{deimel2015novel}, in which authors developed a continuum finger which is pneumatically actuated.

\begin{figure}
	\centering{}\includegraphics[scale=1.58]{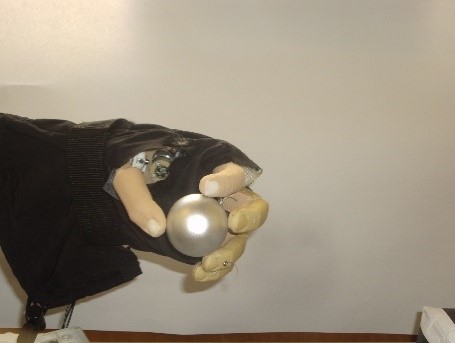}
	\caption{The ISR-Softhand grasping a spherical object. The exerted forces cause undesired deflections on the compliant joint of the thumb, affecting the grasp stability. Grasping heavier objects is impossible, even if the thumb can exert enough normal force}
	\label{fig:1}    
\end{figure}

A major problem associated with compliant joints which are fabricated merely with an elastomer is that they suffer from lateral deflections and twisting. The stiffness of the joint on all axes is dependant to the low Young's modulus of the elastomer. In both of our previous attempts in the development of grasping mechanisms with soft joints, i.e., in Flexirigid \cite{Tavakoli13-1} and ISR-Softhand \cite{Tavakoli2014}, we faced this problem, Figure \ref{fig:1}. This problem was reported also in the development of a soft miniature grasper for surgical applications. The integration of a second material into the joint might be a solution to this problem. This was suggested for instance in \cite{bruyas2014combining}, but in a different context, i.e. for an interventional MRI device. To develop the multi-material joint, authors used a polyjet additive manufacturing unit and combined two materials with a relatively large difference in their Young's modulus. Using the polyjet additive manufacturing it is possible to combine different materials in the same part, which gives a great freedom on design of dual material flexure bearing. This is still an expensive technology and the range of materials available is yet limited.

In the context of prosthetic hands and grasping mechanisms, a multi-material approach was also proposed by Matasuoka's group \cite{xu2011design,deshpande2014development}, where a design for an artificial finger joint for anthropomorphic robotic hands was suggested and tested. In this highly bio-inspired design, each finger is composed of several ABS parts that closely resemble the skeleton of a human finger. It includes ball joints and an elastic capsule around the ball joints. This is a close replica of a human finger which can be used for further studies of such. However, its fabrication and implementation procedure is more labour intensive compared to the other methods such as SDM and HDM.

The goal of our research is to maintain the fabrication simplicity, reduced cost, and at the same time contribute toward two main objectives. The first objective is related with the development of fingers with a better anthropomorphic look and feel. To do so, we proposed a silicone skin to cover the whole finger structure. The current practice of using cosmetics gloves on the prosthetic hands is expensive and it is reported that the cosmetic glove, which covers a hand prosthesis, negatively affects the mechanical efficiency of a prosthesis \cite{smit2013comparison}. The second objective is to enable the design of the finger joints in a way that its stiffness in different axes could be defined within the design. It is desired to establish a trade-off between the desired low stiffness in the flexion axis and the higher stiffness on all other axes of the joints. The former is a very important factor that directly affects the size, weight and power consumption of the actuators. Reducing the required force for flexion of the fingers is one of the most important steps in reducing the overall size and weight of a prosthetic hand.

Our proposed approach and the main novelty of this work for achieving both goals is to decouple the two challenges, i.e., the cosmetic appearance and the stiffness. Another particular novelty of this work is that the desired compliance of the soft fingers is achieved with a rigid matter endoskeleton. This has 3 main advantages. First, to design a finger, one has a broader range of material choices and is not limited only to elastic and soft materials. This is important because 3D printing of arbitrary geometries of rigid materials is currently more affordable and accessible than 3D printing of soft materials. Second, this method allows researchers to fine-tune the properties of the finger joints based on their grasping application, mainly by optimizing the geometry of the endoskeleton and not solely by selection of the joint material. Third, the material for soft skin is decoupled from the material of the joints, while a continuous soft skin is formed. This provides a good freedom for the researchers in choosing the materials and geometry of the soft skin. The soft skin has an important effect on cosmetics and functionality since it provides a larger contact area and a higher friction.

\section{Compliant joints}
\label{sec:2}

Compliant joints and flexural joints have been widely considered in the precision control elements and also in the design of Microelectromechanical systems (MEMS) \cite{sigmund1997design,shih2006two}. When exposed to external forces, the resulting pose of the flexural joints can be precisely modelled and controlled. However, they have not received enough attention in the robotics community, probably because of their non-conventional nature, i.e., due to the difficulties in their design, optimization and fabrication. Among these, the fabrication problem is currently being addressed by additive manufacturing methods.

The topology optimization of the flexural joints concentrated mostly on the objective of distributing a limited amount of material in the design domain such that the output displacement is optimized. This conducts to the maximization of the joint's sensitivity but did not consider the other effects such as the stiffness on other axes, which is an irrelevant subject in that domain due to the generally low external forces. An important factor of flexural joint's is the ratio of $\sigma y/ E$ (yield strength/Young's modulus) \cite{lotti2002novel}. Smaller tensile modulus is beneficial since it reduces the required force for bending of the joints in the desired direction. However, for the material to stay in its elastic zone and not entering the permanent plastic deformation, a high $\sigma y$ is advantageous. 

Lotti et. al. compared the $\sigma y/ E$ of some materials, and based on the results, they developed a single piece finger with Teflon (PTFE) material \cite{lotti2002novel}. The single piece finger was developed by removal manufacturing from a single block. In other research work the same team utilized coil springs as joint of the fingers of the previously mentioned UB-Hand \cite{Lotti05developmentof}. Despite the fact that in the latter case the hand's digits are not anymore composed of a single piece endoskeleton, the concept of utilization of coil springs was interesting due to the low stiffness against the flexion. However, in both cases, the undesired deflection and undesired torsion were not discussed. This aspect was later discussed, in which researchers developed a small compliant grasping mechanism with the shape deposition moulding of compliant materials as joints. In order to reduce the undesired deflections in the fingers they integrated a thin steel reinforcement piece into the joints. In summary, an ideal compliant joint to be used in the digits of a bionic hand or an industrial gripper should be flexible about the flexure axis and have the following characteristics:

\begin{itemize}
	\item Low stiffness in the direction of the desired bending; 
	\item High stiffness against undesired deflection and torsions;
	\item Higher ratio of $\sigma y/ E$ and thus a high elastic property range.
\end{itemize}

In this section, we will discuss several possible versions of the compliant joints and their different properties, as well as their fabrication. This includes joints formed by highly elastic compliant materials, with and without the reinforcements, and joints formed by rigid materials.

\subsection{Compliant joints}
\label{sec:2.A}

\begin{figure}
	\centering{}\includegraphics[scale=0.71]{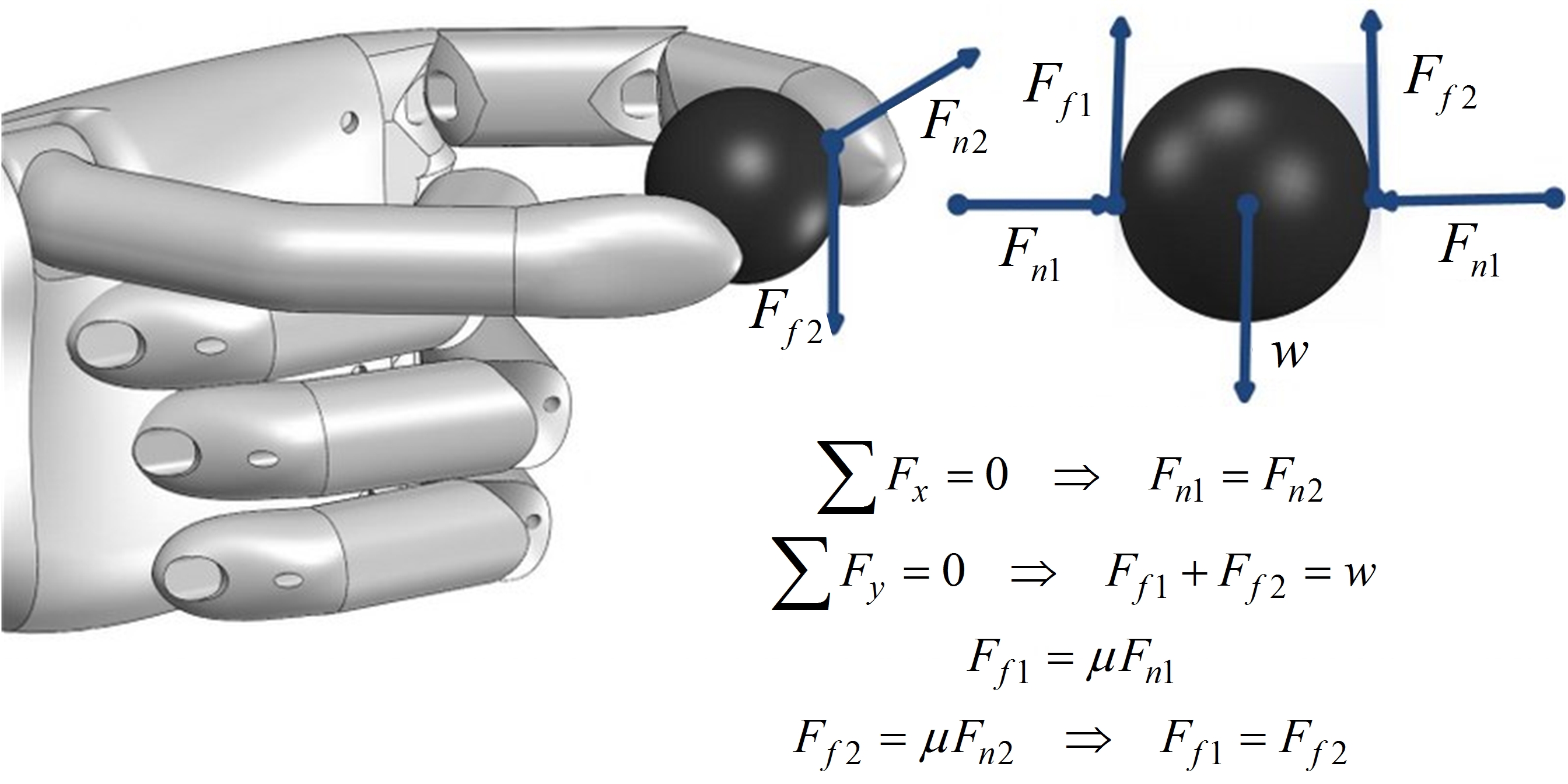}
	\caption{Forces applied to the fingers in a tip pinch}
	\label{fig:2}    
\end{figure}

\begin{figure}
	\centering{}\includegraphics[scale=0.29]{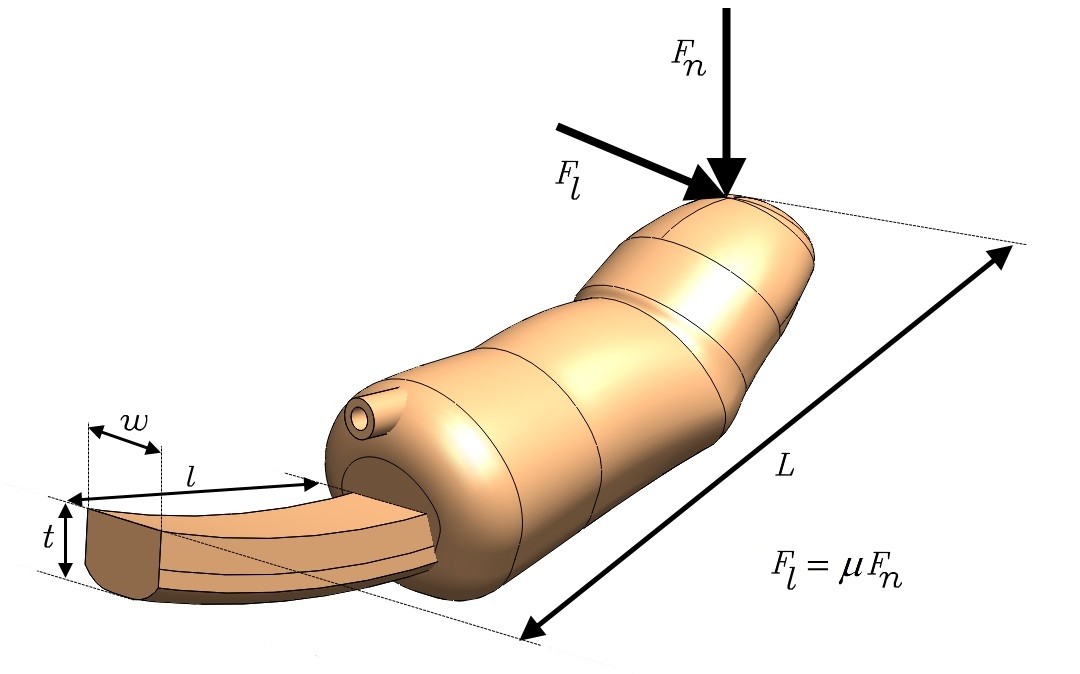}
	\caption{Model of the joint bending}
	\label{fig:3}    
\end{figure}

As discussed, the lateral deflection is the undesired feature of the elastic joints, especially for the thumb, as can be seen in Figure \ref{fig:1}. We can use the beam deflection formula in order to model the joint deflection on both axes, i.e. the desired flexion and the undesired deflection. Figure \ref{fig:2} demonstrates the applied forces to the finger for a tip pinch grasp, where $F_n$ indicates the normal force and $F_f$ indicates the resulting tangential force. Figure \ref{fig:3} demonstrates the normal and lateral forces applied to one finger. The deflection of the joint \cite{gere2008mechanics} can be estimated by:

\begin{equation}
	\delta = \frac{Fl^{3}}{3EI}
\end{equation}
where $\delta$ determines the amount of the deflection of the beam at the length of $l$, which is caused by a force of $F$, $E$ is the material Young's Modulus and $I$ is moment of inertia of the beam. For a rectangular beam $I$ is calculated as:
\begin{equation}
	I= \frac{bh^{3}}{12}
\end{equation}
where $h$ is the dimension in the plane of bending, i.e. in the axis in which the bending moment is applied. Here we call the inertia of the beam in the desired flexion bending as $I_f$ and the inertia of the beam on the undesired lateral bending as $I_l$. Also $\delta_f$ and $\delta_l$ demonstrate the deflection on the flexion axis (desired) and lateral deflection (undesired).
From figure \ref{fig:3}, we have:
\begin{eqnarray}
	I_f=\frac{wt^{3}}{12};
	I_l=\frac{tw^{3}}{12}\\
	\delta_f = \frac{F_fL^{3}}{36wt^3};
	\delta_l= \frac{F_nL^{3}}{36tw^3}
\end{eqnarray}
Thus:
\begin{equation}
	C=\frac{\delta_l}{\delta_f}=\frac{F_fwt^3}{F_ntw^3}=\mu(\frac{t}{w})^2
\end{equation}
where $C$ shows the ratio between the undesired lateral deflection and desired deflection, which should be minimized. At the same time, to have a stable grasp the actual amount of the undesired deflection ($\delta_l$) should be limited to a predetermined value (e.g. less than $1 mm$). By simply increasing the thickness of the joint $t$, we can address both conditions. However, increasing the $t$ results in the cubic growth of the required force for flexion of the joint in the desired direction. Therefore, it is highly desirable to keep the $t$ as low as possible.

\subsection{Reinforced joints}
One method to increase the joint stiffness against the undesired lateral deflection is to integrate into the joint a thin layer of a second material with a higher Young's modulus than the elastic material, and with a high $\frac{w}{t}$ width to thickness value. If this new layer is inserted in the centre of the elastic beam, and considering that its thickness is significantly smaller than the elastic material, the moment of inertia can be calculated as the sum of the moments of the inertia of both beams (for more information please refer to \cite{gere2008mechanics}). That is:

\begin{eqnarray}
	I_{fn}=I_f+\alpha\frac{w_rt_r^{3}}{12}\\
	I_{ln}=I_l+\alpha\frac{t_rw_r^{3}}{12}
\end{eqnarray}
In which  $ I_{fn}$ and $ I_{ln}$ are the new moment of inertia of the joint against the desired flexion and the undesired lateral bending, $w_r$ and $t-r$ are the width and the thickness of the enforcement layer, and $\alpha=frac (Er/Ec)$ is the ratio between the Young's modulus of the integrated reinforcement and the original highly elastic material with low Young's modulus. It should be noted that this model works only if the reinforcement is located in middle of the joint, since the centre of mass stays unchanged in both cases (with or without reinforcement). Otherwise, it becomes necessary to find out the new centre of mass and recalculate the moment of inertia around the new shifted axis.

\begin{figure}
	\centering{}\includegraphics[scale=0.28]{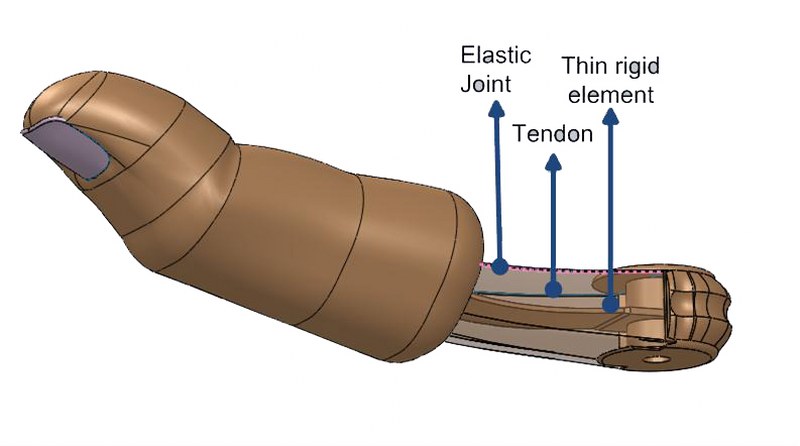}
	\caption{A thin rigid element is integrated to the thumb's flexible joint in order to increase the thumb's lateral stiffness}
	\label{fig:4}    
\end{figure}

\begin{figure*}
	\centering{}\includegraphics[scale=0.82]{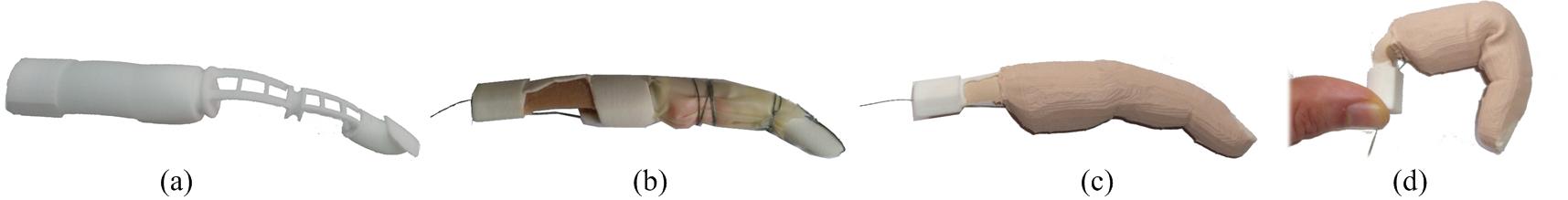}
	\caption{A 3D printed endoskeleton integrating rigid joint reinforcements (a); The DIP joint is filled with some sponge and sealed with an elastic tube (b); This set is then placed in a mould (Figure \ref{fig:6}) which is formed from 3D scanning of a human finger and a resin is cast around the endoskeleton (c) (d).}
	\label{fig:5}    
\end{figure*}

Figure \ref{fig:4} shows a model of a thumb which integrates a thin layer of reinforcement. This was applied and tested in the second version of the ISR-Softhand. Despite effectively reducing the deflections, this model increased substantially the required force for closing the thumb (this will be further analysed in the next section.) For the rest of the fingers, the problem of deflection is less significant than for the thumb, since in grasping of heavier objects, usually all 4 fingers distribute loads among all of them. Therefore, as can be seen in Figure \ref{fig:5}, the reinforcement for the fingers were redesigned and optimized for having less stiffness in the axis of flexion, thus reducing the total power required to close the fingers.

\begin{figure}
	\centering{}\includegraphics[scale=0.36]{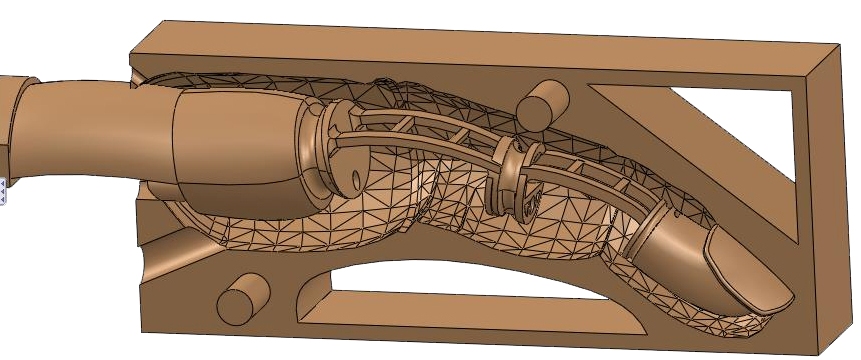}
	\caption{The internal surfaces of the mould are created from the 3D scanning of a human finger}
	\label{fig:6}    
\end{figure}

\section{Flexible Joints through geometrical features on rigid materials}
\label{sec:3}

While the solution of integration of a reinforcement could reduce the deflections, it increased also the joint's stiffness in the flexion direction and the required force for closing of the finger. It is therefore interesting to study alternatives for the rigid reinforcement such as non-rectangular shapes

\subsection{Rigid endoskeleton and the soft skin}
\label{sec:3.A}

It was performed a study which intends to analyse other possible rigid reinforcements. However, we decided to take a different approach. In the previous approach, the elastomer plays the role of the joint. In the new joint, the elastomer would only play the role of the soft skin, which can be designed for better cosmetics with minimal effect on the joint's stiffness. We selected an Ecoflex series silicon (Smooth-on), which is highly stretchable, has a low chemical reactivity, and exhibits very similar mechanical properties to the human skin. Similarities between the mechanical properties of the silicone with human tissues and also the safety of silicone in contact with human body are the reason for being used not only in medical implants, but also in beauty and hair products. For the same reasons, along PDMS, different types of silicone are the most used materials in researchers on soft electronics for wearable technologies that aim to build products in touch with the human skin and human organs. In the specific case of Ecoflex this material is certified for skin touching products.

\begin{figure}
	\centering{}\includegraphics[scale=0.28]{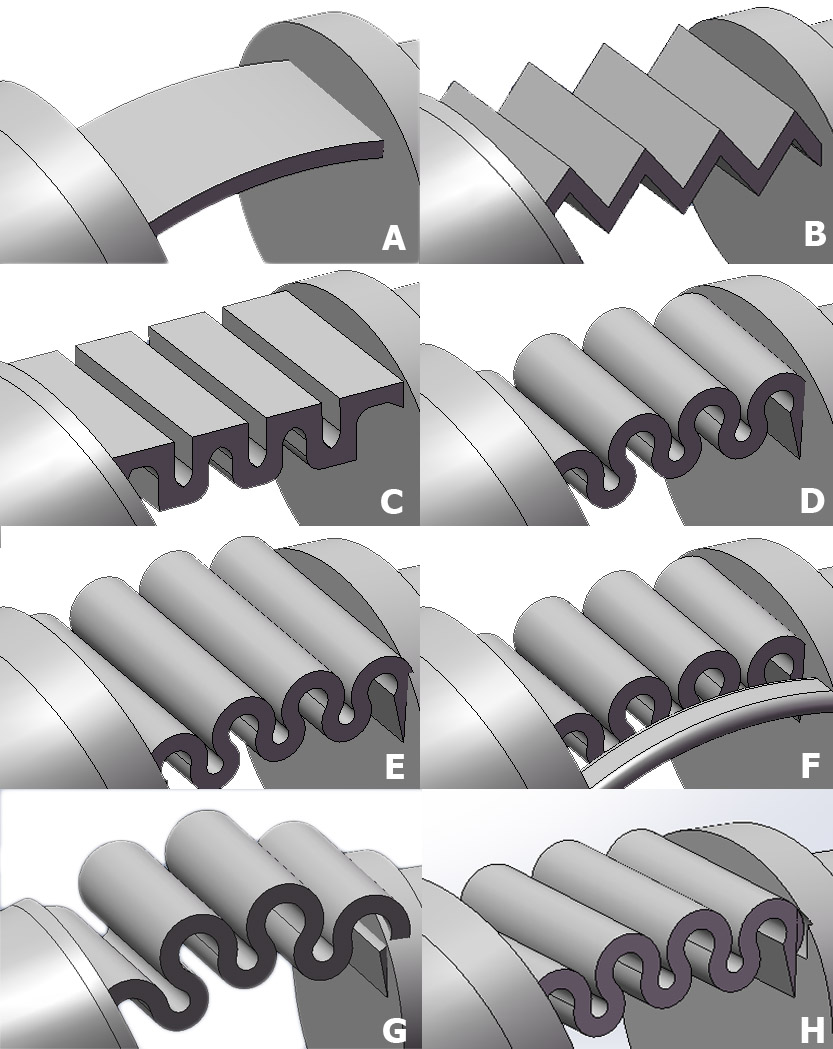}
	\caption{Different joint models used in tests}
	\label{fig:7}    
\end{figure}

\begin{figure}
	\centering{}\includegraphics[scale=0.58]{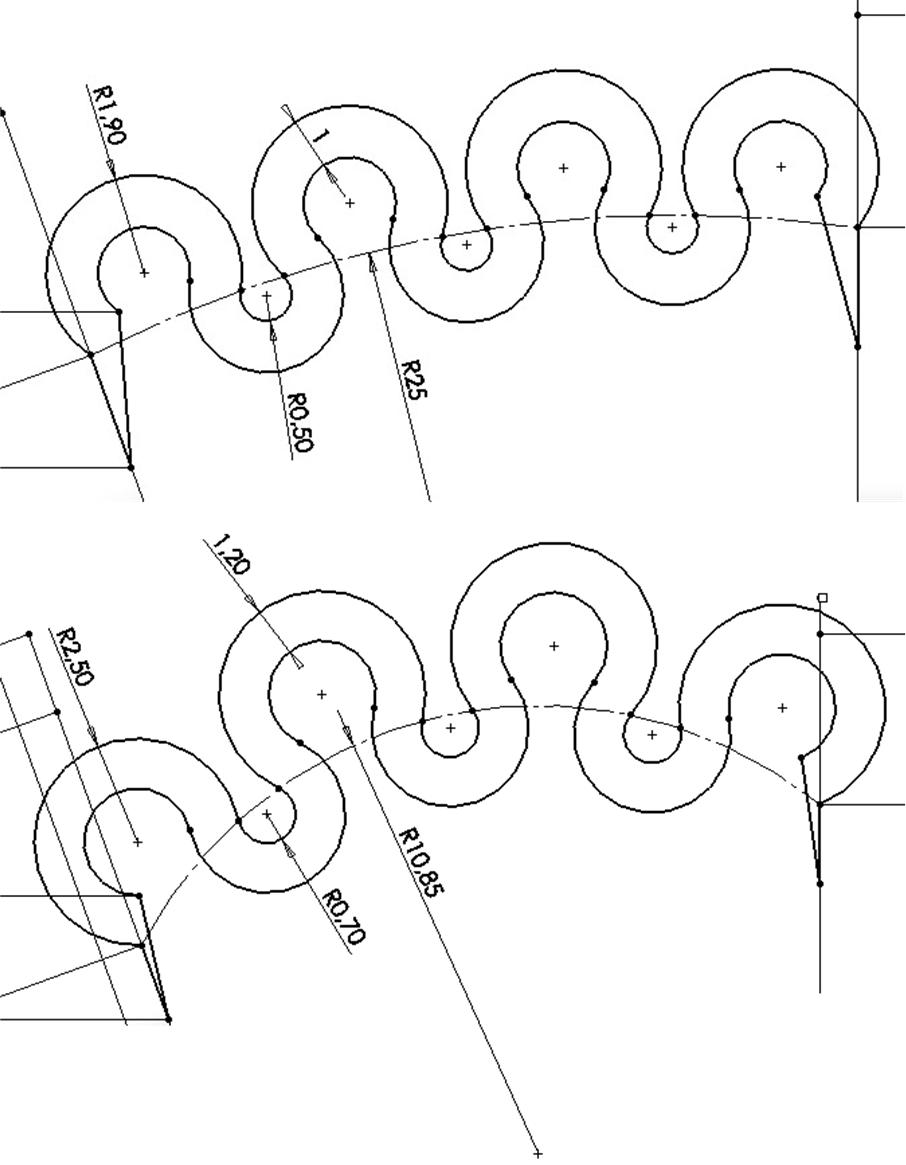}
	\caption{Geometry used in circular pattern for joint samples D, E and H (top), and for sample G (bottom)}
	\label{fig:8}    
\end{figure}

The rigid reinforcement in the previous approach is the actual joint in the new approach. One advantage of the previous elastomer joint was that during the bending, it enables a continuous bending profile which helps in encircling the object. This is an important feature that should be maintained in the new approach. Based on that, the novel joints of the endoskeleton were designed based on repetitive patterns that enable a continuous bending profile. As can be seen in figure \ref{fig:7}, this includes triangular, rectangular and circular patterns in which the final flexion of the joint results from superposing small flexion of each of the patterns. The considered patterns were:

\begin{itemize}
	\item A flat reinforcement, as was already used in the ISR-Softhand (Sample A);
	\item A triangular geometry that is repeated for four times with the width of 13mm and thickness of 1mm (Sample B);
	\item A rectangular geometry that is repeated for four times with fillets of 1mm radius applied at some edges with the width of 13mm and thickness of 1mm (Sample C);
	\item A circular geometry that is repeated for four times with the width of 10mm and thickness of 1mm (Sample D);
	\item Same geometry of Sample D with the width of 13mm and 15mm (Sample H (only for simulation) and Sample E);
	\item An hybrid design with circular geometry as well as flat reinforcements (Sample F);
	\item A circular pattern with a different geometry from Sample D used to analyse the effect of the geometry (Sample G, Figure \ref{fig:8}).
\end{itemize}

\begin{figure}
	\centering{}\includegraphics[scale=0.68]{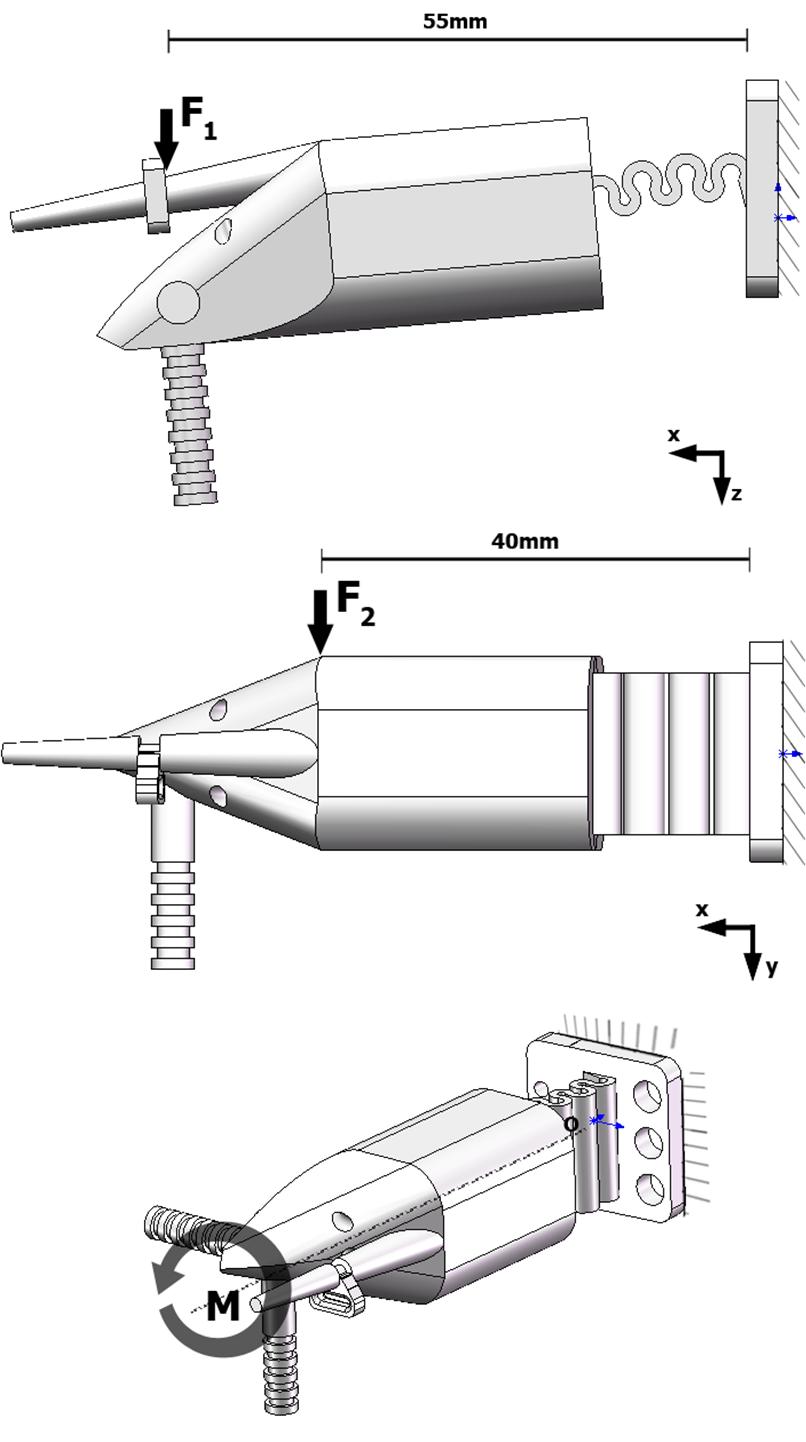}
	\caption{Three types of loads analysed during FEA, maximum bending (top), lateral deflection (middle) and displacement due to torsion (bottom)}
	\label{fig:9}    
\end{figure}

These joints were studied in terms of the desired flexion, undesired lateral deflection and twisting, as well as the stress concentration.  In this way the continuous curve of the finger while bending would be similar to the elastomeric joints. It is difficult to use analytical models to calculate joint's deflection, except for the flat geometry. Therefore a Finite Element Analysis (FEA) was applied to compare several characteristics of the joints. A non-linear solver for large deformations was used in the SolidWorks simulation toolbox. For each joint model we performed the simulations according to the following loads:

\begin{enumerate}
	\item Constant force of 0,8N acting on a Z direction applied at the distance of 0,055m from the fixed base to measure the maximum flexion of all joints for a constant force (Figure \ref{fig:9} top);
	\item Constant force of 4N acting on a Y direction applied at the distance of 0,04m from the fixed base to simulate the lateral deflection (Figure \ref{fig:9} middle);
	\item A constant torque of 0,12 Nm to simulate the twisting of the joint due to the tangential force on the fingertip (Figure \ref{fig:9} bottom).
\end{enumerate}

It should be also mentioned that the absolute value of the forces and torques applied do not affect the conclusions obtained from this research. This is because this is a comparative study and the parameters that will be compared are unitless. For the best functionality of the joints, it is desired to minimize the lateral deflection and also the torsion of the joint due to lateral forces. It is also desirable to have the minimum bending force for flexion of the joint to reduce the size of the required actuators. We therefore created and compared two new criteria:
\begin{enumerate}
	\item LD/FL denotes the amount of the lateral deflection divided by the displacement due to the flexion;
	\item Tor/FL denotes the amount of the displacement due to the torsion divided by the displacement due to the flexion.
\end{enumerate}

Both rates should have a minimum value and the amount of absolute flexion (displacement due to flexion forces) should be also as high as possible.  Furthermore, the FEA analysis allows us to inspect and compare the joints for stress values due to the forces applied on different axes. In all simulations, we considered the properties of the PA2200 material from EOS 3D printers which is used by most of SLS printing service providers. This will be discussed further in the section of materials.

\subsection{Materials and printing method}
\label{sec:3.B}

Before the discussion of the results of the simulation and experiments on different joints, it is important to discuss the role of the materials for the 3D printed joints. The choice of material is important since the achievement of high strains without overcoming the yield strength requires a material with a high $\sigma y/ E$ ratio.  In \cite{lotti2002novel}, the authors calculated this ratio for several plastics and metals, and found out that PTFE has a very good $\sigma y/ E$ ratio of 66.7 compared to 7.1 in Aluminium 7075. Here we made a comparison between the widely used 3D printed methods and materials to find out the best trade-off. 3D printed materials are vast, materials are evolving, and the selection of the best material is not the main focus of this article. For instance, photo-cured polymers can be produced with very different properties. In this study we compare some of the most used materials. 

\begin{figure}
	\centering{}\includegraphics[scale=0.75]{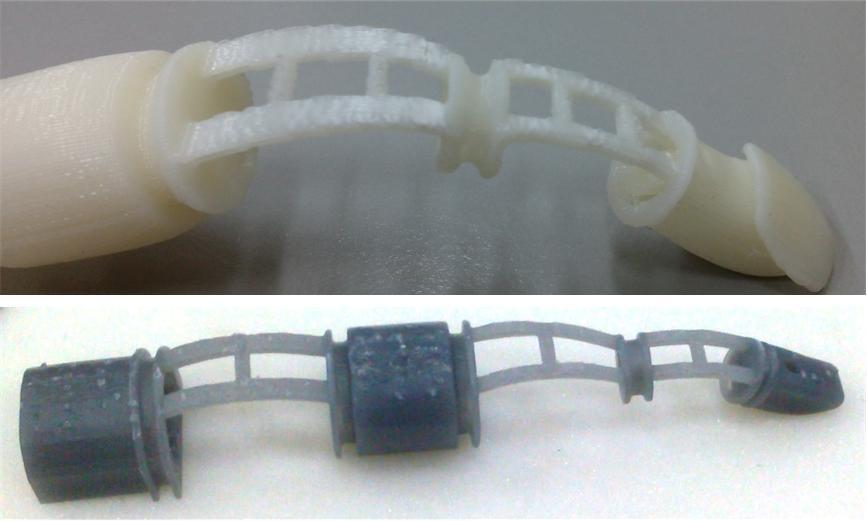}
	\caption{Finger printed with FDM method (top) and SLA method (bottom). In the FDM part the visible defects on the joints reduce the actual yield strength of the joint. The SLA part do not presents the appropriate elastic properties}
	\label{fig:10}    
\end{figure}

Table \ref{tab:1} compares the properties of commonly used 3D printed materials selected from a large 3D printing service provider. As can be seen in Table \ref{tab:1}, a polyamide produced with selective laser sintering (SLS) of a powder has the best $\sigma y/ E$ ratio. In the 3D printing process, not only the material is important, but also the production method makes a difference in the properties of the final prototype, since these properties depend on the density of the object and bonding between the layers. A full comparison of the 3D printed materials is out of scope of this paper. However, considering that the 3D printed parts are generally inferior in properties compared to a machined plastic block, the $\sigma y/ E$ value of the polyamide 3D printed parts seems to be good enough when compared to the list of materials provided in Table \ref{tab:1}. Therefore, we used this material for prototyping of the fingers. We made some tests to reproduce the joints with ABS with a professional grade fused deposition modelling (FDM) 3D printer, as well as an StereoLithography (SLA)  printer, Figure \ref{fig:10}. However, none of these parts were usable in our work. The SLA part presents a relatively low maximum yield stress of the material. On the other hand, the FDM part is low-cost and more available than SLS. However, our tests to reproduce the joints with the FDM printer were unsuccessful. One main problem was that the parts made by FDM are usually with many defects, and these defects are where the joint starts to enter in plastic deformation (some of these effects are visible in Figure \ref{fig:10}). The other problem of the FDM method is that the produced part is often non isotropic, and the tensile strength of the prototype differs in different directions. In both cases, the materials and methods are evolving. However, currently we found out that the SLS method suits better for this application.

\begin{table}
	\caption{Comparison between $\sigma y/ E$ ratio of 3D printed materials \cite{Stratasys, Shapeways}.}
	\label{tab:1}       % Give a unique label
\begin{tabular}[t]{|p{20mm}|p{17mm}|p{17mm}|p{12mm}|}
	\hline
	Material and processing method&Young's modulus $E$ (MPa)&Yield Strength $\sigma y$ (Mpa)&$\sigma y/ E$ x1000\\ \hline
	
	Polyamide PA 2200/SLS& 48& 1700& 28 \\ \hline
	Polyamide PA3200/SLS& 51 & 3200 & 16 \\ \hline
	Resin example/SLA&26&1100&23 \\ \hline
	Alumide/SLS&48&3800&12\\ \hline
	ABS/FDM & 36& 2265& 16 \\ \hline
\end{tabular}
\end{table}

\subsection{Experimental setup}
\label{sec:3.c}

\begin{figure}
	\centering{}\includegraphics[scale=0.48]{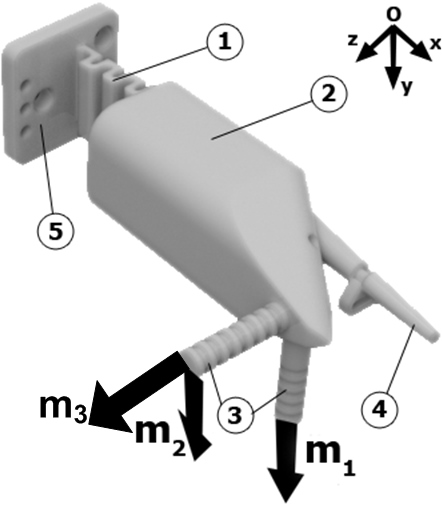}
	\caption{Model of a finger prepared for experimental evaluation. m1 and m2 show the direction of the applied forces for deflection and torsion experiments, respectively}
	\label{fig:11}    
\end{figure}

\begin{figure}
	\centering{}\includegraphics[scale=0.38]{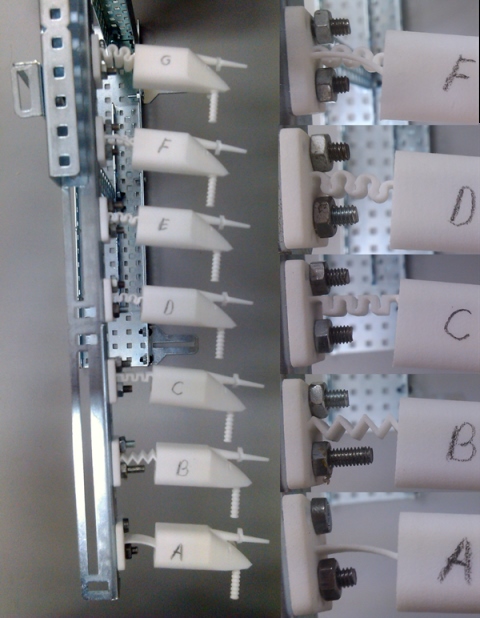}
	\caption{Experimental setup for measuring the undesired deflection}
	\label{fig:12}    
\end{figure}

In order to validate the results of the simulations, we 3D printed some fingers as a test benchmark in order to compare them with the simulation results. As can be seen in Figure \ref{fig:11}, the test model includes a flexible joint (1), a rigid part (2), two perpendicular bars (3) designed to add a mass to simulate the deflection and torsion of the joint in different directions, a sharp tip (4) used for taking the measurements, and a base (5) for fixing the model. In the experimental setup, we could not simulate/visualize the pure torsion as it was simulated in FEA because the added mass would cause a torsion accompanied by a lateral deflection. In any case, this is actually what happens in the actual grasping, i.e. a tangential force is applied to the fingers that cause the deflection and the torsion at the same time. Figure \ref{fig:12} shows the experimental setup for all finger joint models considered in this study.

\section{Results and discussion}
\label{res}

\subsection{FEA simulations}
\label{sec:4a}

A curvature based mesh was used for the FEA simulation. The element size is determined mathematically by the minimum number of elements that fit in a hypothetical circle, while taking into account the user specified minimum and maximum element size. For all models the same values for minimum and maximum element sizes of 0,6mm and 3,0mm were applied. Due to the differences in joint geometries, the number of mesh elements is different for each model falling always in the range between 35000 and 45000 elements.

\begin{figure}
	\centering{}\includegraphics[scale=0.65]{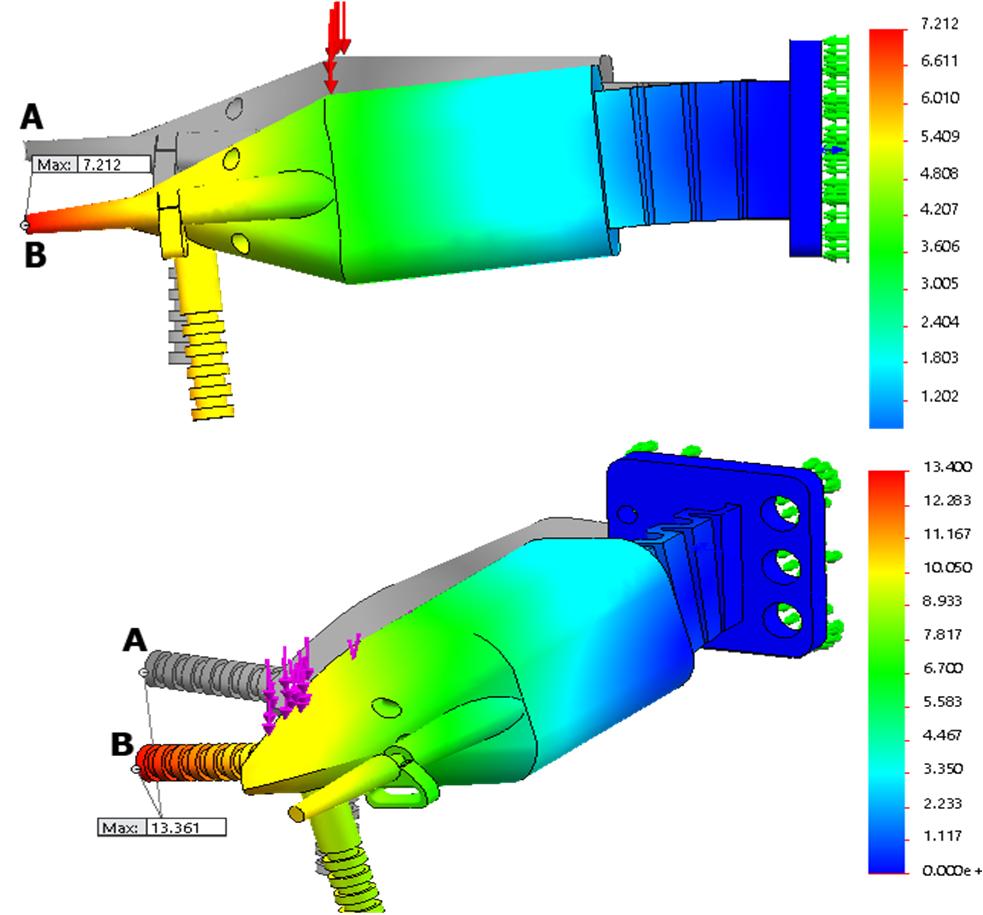}
	\caption{FEA results for lateral displacement (top) and torsion (bottom). A refers to the initial position and B to the final position}
	\label{fig:13}    
\end{figure}

\begin{figure*}
	\centering{}\includegraphics[scale=0.2]{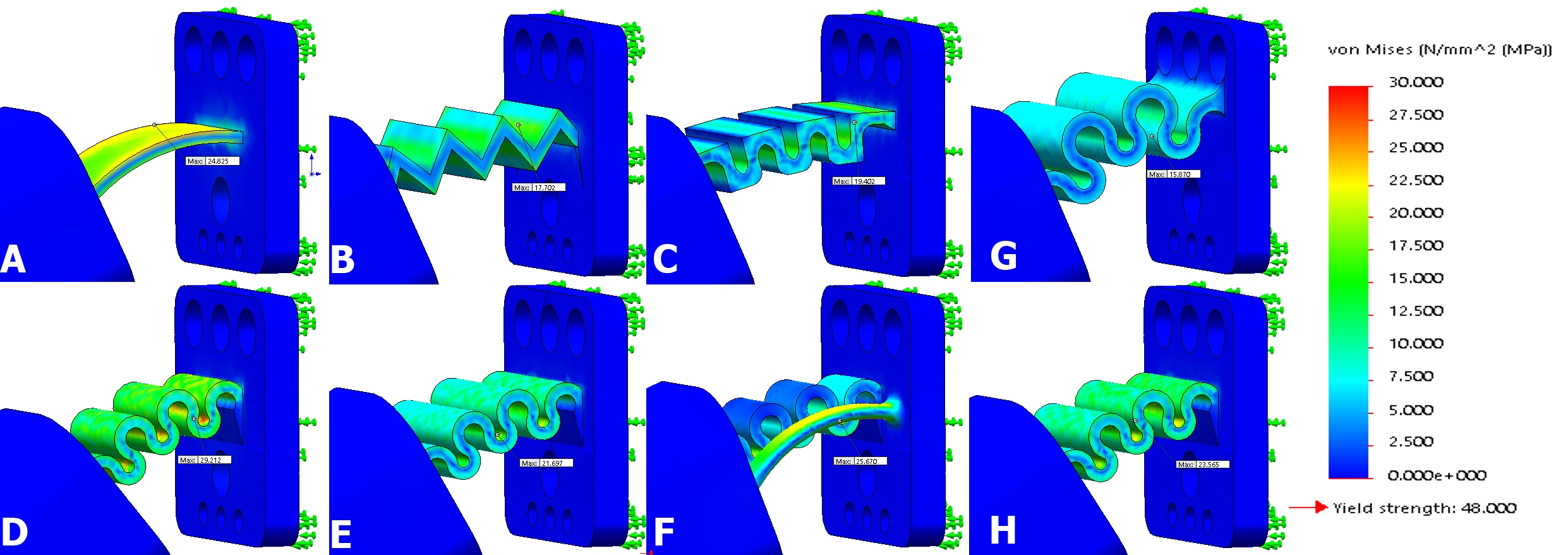}
	\caption{Stress concentrations from FEA}
	\label{fig:14}    
\end{figure*}

Table \ref{tab:2} shows the absolute values for the lateral deflections and the undesired displacement due to twisting for each of the joints. Both values are obtained by FEA and measured on the tip of each finger, Figure \ref{fig:13}. Table \ref{tab:3} shows the absolute value for flexion of each joint based on the FEA. It presents the ratio between the undesired lateral deflection and desired displacement (LD/FL), as well as the ratio between the undesired displacement as a result of twisting (Tor) and the desired flexion (FL). In this study, a minimum value for the ratio LD/FL and Tor/FL is desirable. The stress concentration during the flexion of the joints is depicted in Figure \ref{fig:14}. The maximum stress value that occurs in every joint is shown in Table \ref{tab:4}.

\begin{table}
	\caption{FEA simulation results for the lateral deflection measured after applying a constant force of 4N (F2) and the total displacement on the tip of fingers after applying a torsional torque (M) of 0,12N.m}
	\label{tab:2}       % Give a unique label
	\begin{tabular}[t]{|p{18mm}|p{24mm}|p{27mm}|}
		\hline  Joint model  & Lateral Deflection (mm) & Displacement due to torsion (mm)  \\ 
		\hline A  & 2,3 & 46,0   \\ 
		\hline B  & 3,8 & 26,5  \\ 
		\hline C  & 5,0 & 13,4 \\ 
		\hline D  & 21,7 & 38,6  \\ 
		\hline E  & 7,6 & 19,4  \\
		\hline F  & 1,1 & 25,7 \\
		\hline G  & 9,1 & 17,4  \\ 
		\hline H  & 11,2 & 22,4  \\
		\hline 
	\end{tabular}
\end{table}

\begin{table}
	\caption{FEA simulation results for the ratio LD/FL and the ratio Tor/FL. displacement.}
	\label{tab:3}       % Give a unique label
	\begin{tabular}[t]{|p{18mm}|p{20mm}|p{13mm}|p{13mm}|}
		\hline Joint model & Flexion (mm) & LD/FL & Tor/FL\\ 
		\hline A  & 22,5 & 0,10 & 1,07 \\ 
		\hline B  & 23,4 & 0,17 & 1,19 \\ 
		\hline C  & 21,2 & 0,23 & 0,63 \\ 
		\hline D  & 37,3 & 0,58 & 1,04 \\ 
		\hline E  & 27,6 & 0,28 & 0,70 \\
		\hline F  & 18,2 & 0,06 & 1,07 \\
		\hline G  & 21,2 & 0,43 & 0,82 \\ 
		\hline H  & 30,9 & 0,36 & 0,82 \\ 
		\hline 
	\end{tabular} 
\end{table}

\begin{table}
	\caption{FEA simulation results for the maximum stress value on each of the joints as a result of an equal load on each joint for flexion and lateral deflection}
	\label{tab:4}       % Give a unique label
	\begin{tabular}[t]{|p{18mm}|p{24mm}|p{27mm}|}
		\hline Joint model   & Flexion (MPa) & Lateral Deflection (MPa)  \\ 
		\hline A  & 24,8 & 18,9   \\ 
		\hline B  & 17,6 & 13,8   \\ 
		\hline C  & 18,4 & 54,5   \\ 
		\hline D  & 29,2 & 95,7   \\ 
		\hline E  & 21,7 & 52,5   \\ 
		\hline F  & 25,7 & 25,1   \\ 
		\hline G  & 15,8 & 42,0   \\ 
		\hline H  & 23,6 & 68,1   \\ 
		\hline 
	\end{tabular} 
\end{table}

\subsection{Experimental tests}
\label{sec:4b}
Table \ref{tab:5} shows the results of the undesired deflections for two different masses of $150g$ and $300g$ applied in m1 and m2, Figure \ref{fig:11}. To test the flexion of each joint under the effect of a constant force we measured the desired displacement with the masses of $40g$ and $80g$. Table \ref{tab:6} shows the amount of the flexion on each of the joints when a mass of $40g$ and $80g$ is applied in the direction of m3, Figure \ref{fig:11}. Here, higher displacement is desired since this means that the joint offers less resistance against the desired flexion. It is also desired to minimize LD/FL and Tor/FL, and to maximize the displacement due to flexion for a constant force.  The values of LD/FL and Tor/FL for the experimental tests are also presented in Table \ref{tab:6}.

\begin{table*}
	\begin{center}
	\caption{Experimental results for undesirable deflection}
	\label{tab:5}       % Give a unique label
	\begin{tabular}{|p{17mm}|p{17mm}|p{17mm}|p{20mm}|p{17mm}|p{20mm}|}
		\hline
	Joint model	&     Parameter      &     m1.1= 150g     &         m2.1= 150g          &     m1.2=300g      & m2.2=300g \\ \hline
		&           & Lateral deflection & Displacement due to torsion & Lateral deflection &  Displacement due to torsion  \\ \hline
		A &   y, mm   &        0,0         &             4,8             &        0,0         &   18,3    \\ \hline
		&   z, mm   &        1,2         &             2,4             &        3,8         &   13,0    \\ \hline
		& delta, mm &        1,2         &             5,4             &        3,8         &   22,5    \\ \hline
		B &   y, mm   &        0,0         &             2,9             &        0,0         &    7,3    \\ \hline
		&   z, mm   &        5,2         &             6,4             &        11,3        &   15,0    \\ \hline
		& delta, mm &        5,2         &             7,0             &        11,3        &   16,6    \\ \hline
		C &   y, mm   &        0,0         &            -2,2             &        0,0         &    2,2    \\ \hline
		&   z, mm   &        5,9         &             5,8             &        12,3        &   12,5    \\ \hline
		& delta, mm &        5,9         &             6,2             &        12,3        &   12,6    \\ \hline
		D &   y, mm   &        0,0         &            -6,7             &        0,0         &    8,9    \\ \hline
		&   z, mm   &        20,0        &            19,7             &        32,0        &   32,9    \\ \hline
		& delta, mm &        20,0        &            20,8             &        32,0        &   33,5    \\ \hline
		E &   y, mm   &        0,0         &             1,5             &        0,0         &   14,9    \\ \hline
		&   z, mm   &        9,6         &             9,0             &        15,3        &    2,1    \\ \hline
		& delta, mm &        9,6         &             9,1             &        15,3        &   14,1    \\ \hline
		F &   y, mm   &        0,0         &             3,0             &        0,0        &   22,0    \\ \hline
		&   z, mm   &        1,8         &             2,7             &        2,4         &    9,4    \\ \hline
		& delta, mm &        1,8         &             4,0             &        2,4        &   24,0    \\ \hline
		G &   y, mm   &        0,0         &             2,7             &        0,0         &    5,4    \\ \hline
		&   z, mm   &        10,4        &             9,4             &        18,4        &   18,2    \\ \hline
		& delta, mm &        10,4        &             9,8             &        18,4        &   18,9    \\ \hline
	\end{tabular}
\end{center}  
\end{table*}

\begin{table*}
	\begin{center}
	\caption{Joints flection, LD/FL and Tor/FL}
	\label{tab:6}       % Give a unique label
	\begin{tabular}{|c|c|c|c|c|c|}
		\hline
	Joint model	&     Parameter         & \multicolumn{2}{c|} {Flexion} & LD/FL & Tor/FL \\ \hline
		&           & m3.1= 40g &     m3.2=80g      &         &  \\ \hline
		A &   x, mm   &     8     &        4,8        &         &  \\ \hline
		&   z, mm   &   10,8    &       18,2        &         &  \\ \hline
		& delta, mm &   13,4    &       23,6        &  0,16   &   0,85   \\ \hline
		B &   y, mm   &   10,4    &       20,2        &         &  \\ \hline
		&   z, mm   &   12,1    &       18,3        &         &  \\ \hline
		& delta, mm &   16,0    &       27,3        &  0,41   &   0,56   \\ \hline
		C &   y, mm   &    8,7    &       17,3        &         &  \\ \hline
		&   z, mm   &   14,6    &       23,5        &         &  \\ \hline
		& delta, mm &   17,0    &       29,2        &  0,42   &   0,40   \\ \hline
		D &   y, mm   &   17,2    &       27,5        &         &  \\ \hline
		&   z, mm   &   15,3    &       19,1        &         &  \\ \hline
		& delta, mm &   23,0    &       33,5        &  0,96   &   0,95   \\ \hline
		E &   y, mm   &   10,4    &       20,8        &         &  \\ \hline
		&   z, mm   &   14,9    &       22,0        &         &  \\ \hline
		& delta, mm &   18,2    &       30,3        &  0,50   &   0,44   \\ \hline
		F &   y, mm   &    7,6    &       12,6        &         &  \\ \hline
		&   z, mm   &   11,2    &       18,0        &         &  \\ \hline
		& delta, mm &   13,5    &       22,0        &  0,11   &   0,97   \\ \hline
		G &   y, mm   &    7,6    &       14,8        &         &  \\ \hline
		&   z, mm   &   13,4    &       20,1        &         &  \\ \hline
		& delta, mm &   15,4    &       25,0        &  0,74   &   0,71   \\ \hline
	\end{tabular} 
\end{center}
\end{table*}

\subsection{Discussion}
\label{sec:4c}

\begin{figure}
	\centering{}\includegraphics[scale=0.47]{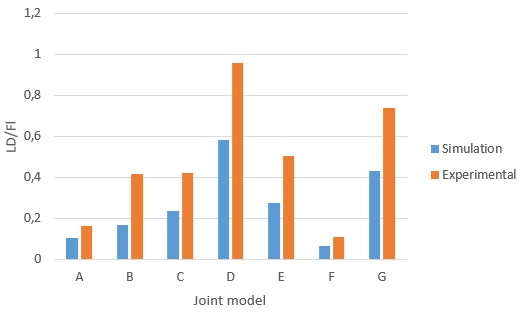}
	\caption{Comparison of LD/FL obtained from experimental tests and FEA simulation}
	\label{fig:15}    
\end{figure}

Analysing Table \ref{tab:3} with results from simulations and Table \ref{tab:5} with results from experimental tests, it can be concluded that joint F expresses the overall best LD/FL ratio, Figure \ref{fig:15}. However, the amount of flexion that it can achieve is lower than all of the other joints. This is due to the two lateral supports parallel to the circular joint, which limits the lateral deflections, but also increases the stiffness against the desired flexion. Therefore, it does not reflect a good trade-off. On the other hand, joint model D offers the highest amount of flexion among all joints, both in simulation and experimental results. However, results of the stress analysis reflected on table \ref{tab:4} show that it also should bear the highest stress concentration compared to other models. As can be seen in Table \ref{tab:6}, where experimental results are reflected, joint model C presents the best Tor/FL ratio.  This is closely followed by Joint B and E. However, joint B does not reflect the relatively good characteristics regarding the torsional stiffness. Looking at the simulation results of the stress analysis, Table \ref{tab:4} and Figure \ref{fig:14}, we can see that joints C and E are also very similar in terms of the stress concentration. Both of these joints (C and E) express good trade-offs. In this case joint C is slightly a better trade-off for the first objective (minimize undesired deflections - e.g. for the thumb) and joint E is a slightly better trade-off toward the second objective (minimize joints stiffness against the desired flexion - e.g. for other digits).

\subsection{Hand’s digits}
\label{sec:4d}

As it was discussed earlier, Ecoflex-30 (smooth-on) was used as the soft skin that covers the endoskeleton. The low Young's modulus of Ecoflex has a minimal effect on the overall stiffness of the joints. In this way, one can optimize further the geometrical features for the best properties, without concerning much about the effect of the enveloping polymer. The enveloping polymer keeps its other important roles on formation of larger contact areas, a high friction coefficient, and the desired cosmetics.

\begin{figure}
	\centering{}\includegraphics[scale=0.23]{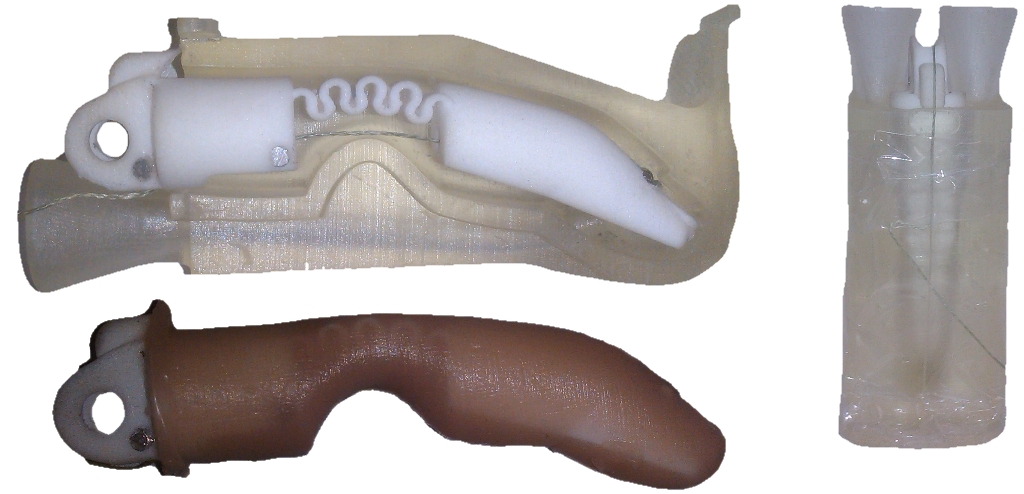}
	\caption{Fabrication process of the finger. The prepared 3D printed endoskeleton is placed into the mould and filled with silicone resin}
	\label{fig:16}    
\end{figure}

Figure \ref{fig:16} shows the fabrication method of the fingers with the novel approach. In contrary to the previous method, Figure \ref{fig:5}, here we do not fill the joint with the sponge prior to casting. Since the enveloping polymer benefits from a low Young's modulus, it can fill the joint without having a significant effect on the joint behaviour. A recent study on the cosmetics gloves for prosthetics hands considered that silicone gloves are more suitable for application on articulating hand prosthesis, as they had a lower joint stiffness and required a lower maximum joint torque \cite{smit2013comparison}.

\begin{figure}
	\centering{}\includegraphics[scale=0.25]{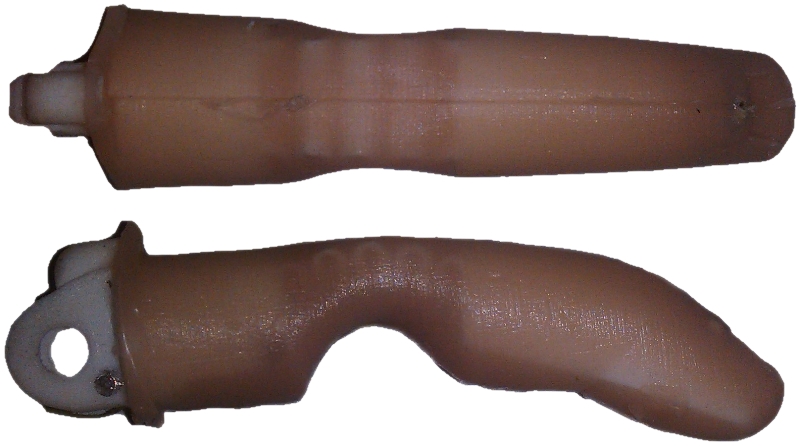}
	\caption{A sample prototype of the finger}
	\label{fig:17}    
\end{figure}

Compared to the previous approaches, the new method is significantly simpler and less labour intensive, resulting in a more uniform skin that fully envelops the endoskeleton. In this way a more anthropomorphic look and feel is achieved. Figure \ref{fig:17} shows a sample prototype of the fingers. As can be seen in Figure \ref{fig:16} and Figure \ref{fig:17}, the mould shape is designed to minimize the amount of the silicone around the endoskeleton. The reason for this specific shape is to avoid the buckling effect of the silicone, which does not allow a full flexion of the joint.

Based on our previous tests with the ISR-Softhand \cite{Tavakoli2014}, we found out that the thumb should have the highest stiffness against lateral deflection and undesired twisting, since it is suffering higher tangential and normal forces than the other fingers. That is, to grasp a heavy object the thumb should oppose the force that on the other side is divided by the other 4 fingers. For the other fingers such stiffness can be lower, in favour of lower required force for full flexion of the fingers. This is important to reduce the required force for closing the fingers, thus reducing the size and weight of the actuators on a prosthetic terminal. Therefore, joint C is the preferred configuration for the thumb and joint E is the preferred configuration for the other fingers.

\begin{figure}
	\centering{}\includegraphics[scale=0.68]{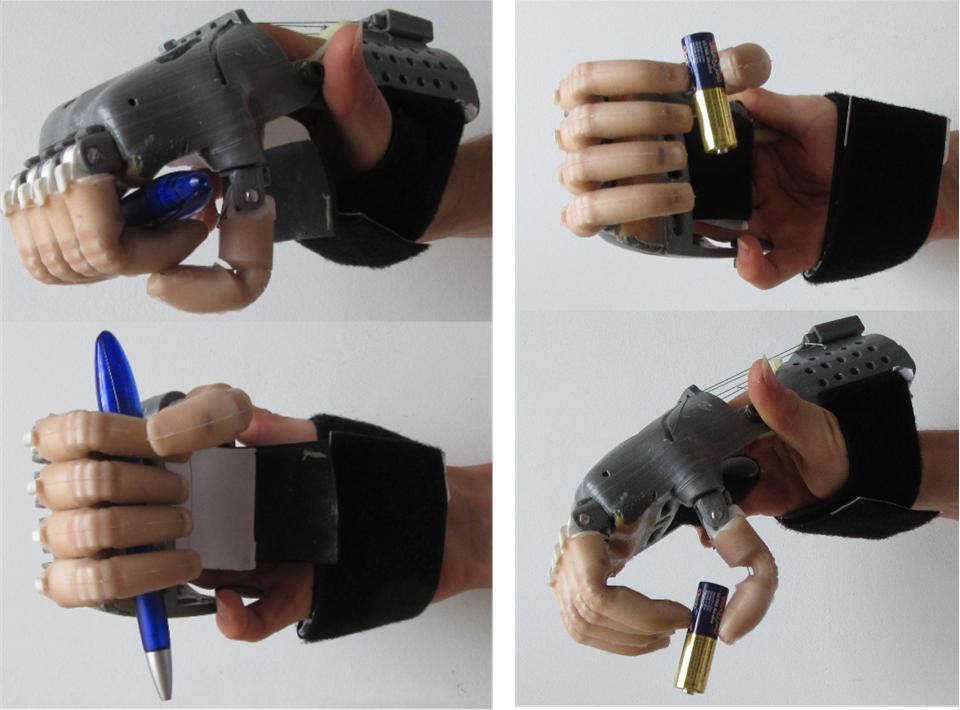}
	\caption{A prosthetic hand developed for an 8 years old amputee using the fingers with the developed compliant joint. It is based on the E-nable community designs (actuated by the wrist/elbow movement)}
	\label{fig:18}    
\end{figure}

Figure \ref{fig:18} shows an example of application of these joints on a body-actuated prosthetic hand which was developed for an 8 year old amputee based on the E-nable community concept. The thumb was selected from the joint type C (rectangular) and the other 4 fingers were selected from the joint type D (circular). The dimensions of the joints were then adjusted based on the size of the hand.

\subsection{Integrated finger comparison}
\label{sec:4e}

\begin{figure}
	\centering{}\includegraphics[scale=0.74]{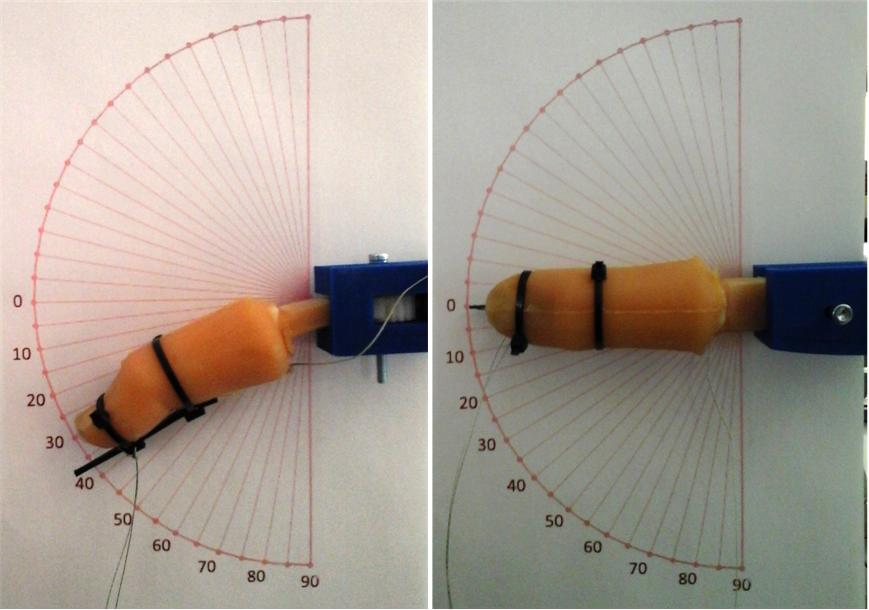}
	\caption{The setup for measuring the flexion force (left) and the undesired lateral torsion (right) of the two versions of the thumb}
	\label{fig:19}    
\end{figure}

A sample thumb digit was developed by this method with the same size of the previous thumb based on elastomeric joint in the ISR-Softhand. On the ISR-Softhand the joint is made with an elastomer (PMC-780-smooth-on). In the new version, an endoskeleton with a rectangular geometry was embedded in a silicone skin. Both thumbs were then compared in terms of required force for flexion and torsion. Figure \ref{fig:19} shows the setup used for measuring the flexion and torsion. For flexion, the tendon is pulled by a cable and the bending angle at the tip of the finger is measured. For torsion, a tangential force is applied in the fingertip and the bending angle is measured. In both cases the force is measured by a digital force gauge. 

\begin{figure}
	\centering{}\includegraphics[scale=0.53]{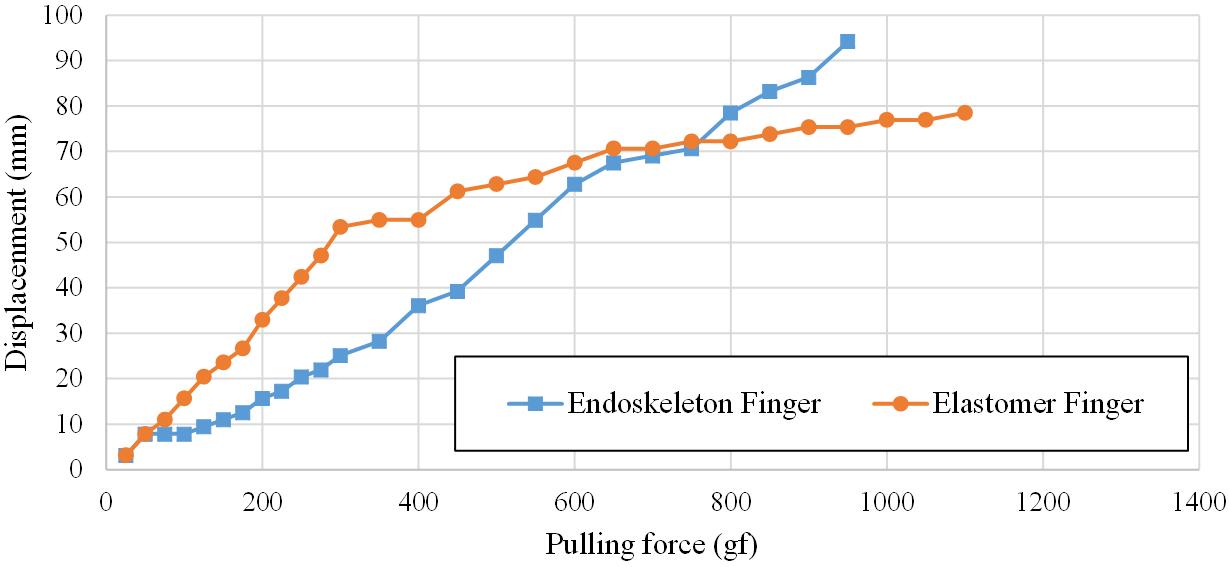}
	\caption{Angular displacement of the two types of fingers against the pulling force applied to the tendon}
	\label{fig:20}    
\end{figure}

Figure \ref{fig:20} shows the comparison between the two fingers on flexion. The endoskeleton design in this finger is rectangular which is specially selected for the thumb. It is not the best option in terms of bending, but it is better than the circular geometry in terms of lateral deflection. The endoskeleton finger covered by the silicone requires a higher force in the beginning of the bending range and a lower force at the end of the bending range. Indeed, the elastomer finger has a non-linear behaviour, and reaches to a threshold around 60 degrees, after which additional bending of the joint becomes very insensitive to the increase of the force. The reason for this is that the elastomer's bending has a tension component on the upper layer of the elastomer, and a compression component in the lower layer of the joint, and after this threshold more compression cannot be achieved. This is also referred as the “buckling effect”, as this happens in the lower half of the elastomeric joint

\begin{figure}
	\centering{}\includegraphics[scale=0.54]{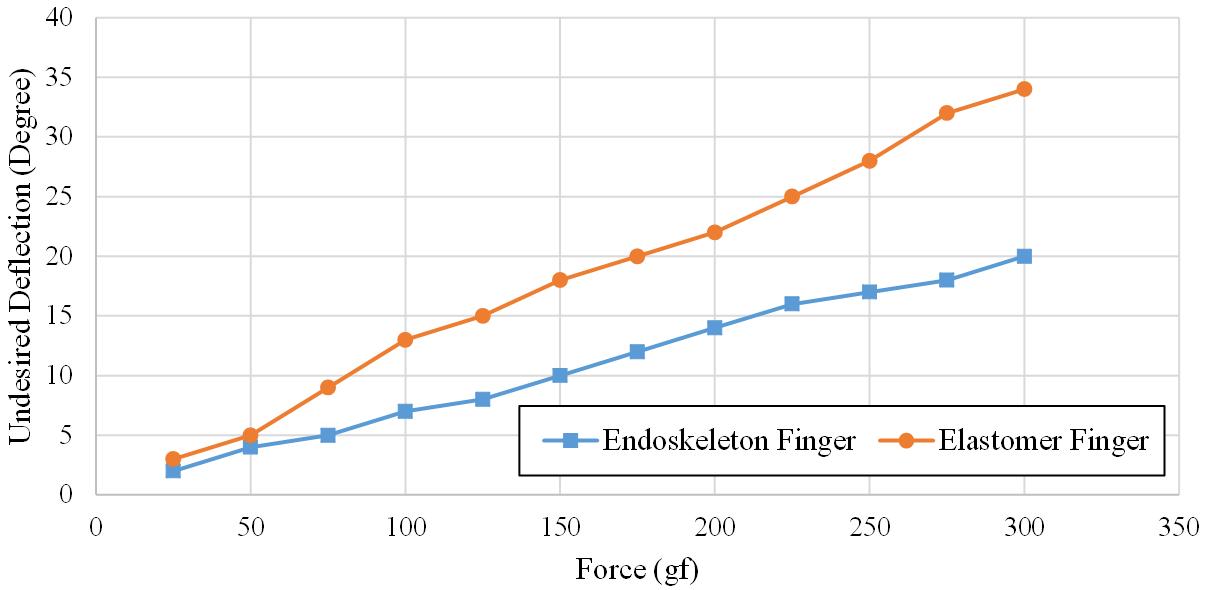}
	\caption{Undesired angular displacement due to torsion for the two types of fingers. The load is applied tangential to the finger, on its tip}
	\label{fig:21}    
\end{figure}

Figure \ref{fig:21} shows the undesired deflection of each of the fingers, as a result of a tangential force applied at the finger tip. Such force resembles the undesired deflection and torsion at the same time, and also resembles the force which is being applied to the finger during the grasping. As can be seen, the endoskeleton joint behaves considerably better than the PMC elastomeric joint.

\section{Conclusion}
\label{sec:sss1}

In this article we introduced a novel method for fabrication of soft fingers for grasping applications. Compared to the previous methods of fabrication of soft fingers, this method presents several advantages. This includes easier fabrication (e.g. one step casting), more anthropomorphic look and a more uniform soft skin cover for the endoskeleton. In addition, this approach addresses an important problem in the fabrication of soft fingers. That is, by optimization of the endoskeleton's geometry, the undesired lateral deflections and the undesired twisting are minimized, and furthermore the joint's stiffness against the desired flexion can be reduced. The latter is especially important, since it results in reducing the size of the actuators which contributes to reduction of the size and weight of the prosthetic terminals. We also compared several geometries of the flexural joints both by simulation and experimental results, suggesting that a rectangular geometry for the thumb and a circular geometry for the other fingers are preferred.

\section*{Acknowledgments}
This research work was partially supported by the Portuguese Foundation of Science and Technology, the Carnegie Mellon-Portugal under contracts CMUP- EPB/ TIC/ 0036/ 2013 and CMUP-ERI/TIC/0021/2014, and by Portugal 2020 project POCI-01-0145-FEDER-016418 by UE/FEDER through the program COMPETE2020.

%Bibliography
\bibliographystyle{unsrt}  
\bibliography{templateArxiv}

\end{document}